LETTER

# Fast Density Codes for Image Data

Pierre Courrieu

Laboratoire de Psychologie Cognitive, UMR CNRS 6146, Université de Provence – Centre St Charles
Bat. 9, Case D, 3 place Victor Hugo, 13331 Marseille cedex 3, France
E-mail: Pierre.Courrieu@univ-provence.fr

(Submitted on October 22, 2007)

*Abstract*—Recently, a new method for encoding data sets in the form of "Density Codes" was proposed in the literature (Courrieu, 2006). This method allows to compare sets of points belonging to every multidimensional space, and to build shape spaces invariant to a wide variety of affine and non-affine transformations. However, this general method does not take advantage of the special properties of image data, resulting in a quite slow encoding process that makes this tool practically unusable for processing large image databases with conventional computers. This paper proposes a very simple variant of the density code method that directly works on the image function, which is thousands times faster than the original Parzen window based method, without loss of its useful properties.

## 1. Introduction

Recently, a new method for encoding sets of points belonging to multidimensional spaces has been proposed by [1]. Given a set of data points, this method builds a deterministic sequence of points, called "density code", whose spatial distribution approximates that of the data. However, contrarily to data points, code points are strictly ordered in a non-arbitrary way, which makes any pair of density codes comparable. This allows building powerful code dissimilarity functions that can be simply made invariant to a wide variety of affine and non-affine natural transformations. As demonstrated in [1], this is clearly a promising approach to pattern recognition problems. However, it turned out that the encoding process is quite slow (several minutes for a 200×200 pixels image), mainly due to the use of a Parzen window scheme to smoothly approximate the data density [2], and to the requirement of integrating kernel functions. It was also proposed, in [1], a parallel neuron-like implementation of the method for image data, which could be fast when actually running on a parallel architecture. Unfortunately, the simulation of this parallel implementation on a conventional sequential computer is much slower than the basic implementation, that is itself too slow to be used on-line, or to process large image databases and to perform realistic simulations of neural processing models. Given that most simulations are performed on conventional computers, there is clearly a need to find a rapid way of building density codes for common data types such as images or image sequences, without loosing the useful properties of these codes. Fortunately, this is easy if one takes into account the "array of pixels" structure of image data, as we shall see.

## 2. Theory and Formulation

*Background*—First, we rapidly summarize the density code method foundations as stated in [1]. Let $f(X)$ be a probability density function on $R^n$, then the density of the $(n-k+1)$-dimensional marginal variable $(x_k, x_{k+1},...,x_n)$ is given by:

$$f(\bullet,...,\bullet,x_k,x_{k+1},...,x_n) = \int_{R^{k-1}} f(x_1,...,x_n)\, dx_1...dx_{k-1}.$$

The density of the one-dimensional conditional variable $(x_k \mid x_{k+1} = a_{k+1},...,x_n = a_n)$ is given by:

$$f(x_k | x_{k+1} = a_{k+1},...,x_n = a_n) = \frac{f(\bullet,...,\bullet,x_k,a_{k+1},...,a_n)}{f(\bullet,...,\bullet,\bullet,a_{k+1},...,a_n)}.$$

The cumulative probability function of this variable is given by:

$$\Pr(x_k \le b | x_{k+1} = a_{k+1},...,x_n = a_n) = \int_{-\infty}^{b} \frac{f(\bullet,...,\bullet,x_k,a_{k+1},...,a_n)}{f(\bullet,...,\bullet,\bullet,a_{k+1},...,a_n)} dx_k.$$

In compliance with [1], one simplifies the above notation by:

$$\Pr(x_k \le b | x_{k+1} = a_{k+1},...,x_n = a_n) = F(b | a_{k+1},...,a_n),$$

where the uppercase $F$ recalls that de corresponding density function is $f$.

For a random variable $X \in R^n$ with density $f$, one defines a mapping $P[f]$ from $R^n$ to $(0,1)^n$ as:

$$P[f](X) = (P_1[f](X), P_2[f](X),...,P_n[f](X)),$$

where

$$P_n[f](X) = F(x_n) = \int_{-\infty}^{x_n} f(\bullet,...,\bullet,t)\, dt,$$

and

$$P_k[f](X) = F(x_k | x_{k+1},...,x_n),\quad 1 \le k \le n-1.$$

Let $U$ be a random variable uniformly distributed in $(0,1)^n$, and assume that $P[f]$ is a bijection, then, according to Theorem 1 from [1], the reciprocal bijection $P^{-1}[f]$ has the following property:

$P^{-1}[f](U)$ is distributed as $X$, with a probability density equal to $f$.

The above result is the foundation of the density coding method, since a density code is simply a realization of the mapping $P^{-1}[f](U)$ with a fixed sequence of $m$ distinct values of $U$. Theorem 1 from [1] also states that a sufficient condition for $P[f]$ to be a bijection is that $f$ be continuous and nowhere zero. This fact motivated the use in [1] of a superposition of continuous kernel functions centered on data points and asymptotically decreasing to zero, in order to approximate $f$. This solution works, however it is computationally slow.

Let $Y \in R^n$ be a random variable functionally related to $X$ by a continuous invertible transformation $\psi$, then:

$$Y = \psi(X),\text{ and } g(Y) = f(X)J^{-1}(\psi(X)),$$

where $g$ is the probability density of $Y$, and $J(\psi(X))$ is the Jacobian determinant of the transformation. Theorem 2 from [1] states that if the Jacobian matrix $((\partial \psi_i / \partial x_j)(X))$ of the transformation is everywhere triangular ($j < i \Rightarrow \partial \psi_i / \partial x_j = 0$), and has a strictly positive diagonal ($\partial \psi_i / \partial x_i > 0,\ 1 \le i \le n$), then:

$$P^{-1}[g](U) = \psi(P^{-1}[f](U)).$$

The above result is the foundation of the density code comparison method. The set of transformations that have the required properties includes a wide variety of affine and non-affine natural transformations. However, certain common affine transformations such as rotations and reflections are excluded.

***Encoding Algorithm***—Now, how can we translate the above model for finite array data types such as images or sequences of images? First, we assume that the (unknown) original image function $h$ is a positive function whose continuous support is a hyper-rectangle $R_h = [0, S_1] \times ... \times [0, S_n]$, where the $S_i$'s are expressed in pixel side units, and $h = 0$ outside this hyper-rectangle. The image discretization results in a (given) multidimensional array $\tilde{h}$ where each (hyper-) pixel, with integer coordinates $(x_1,...,x_n)$, $1 \le x_i \le S_i$, has the mean value of $h$ on the unit volume hypercube $[x_1 - 1, x_1] \times ... \times [x_n - 1, x_n]$, that is:

$$\tilde{h}(x_1,...,x_n) = \int_{x_1-1}^{x_1} ... \int_{x_n-1}^{x_n} h(t_1,...,t_n)\, dt_1...dt_n.$$

This simple and quite reasonable approximation of the discretization process allows us to reduce all subsequent integrals to finite discrete sums of pixel values. However, image data are subject to variations of foreground and background lighting that are irrelevant for shape recognition. An image affine transformation allows us to partly solve this problem. Let $\min(\tilde{h})$ and $\max(\tilde{h})$ denote respectively the minimum and maximum pixel values in $\tilde{h}$, then one transforms the array $\tilde{h}$ into an array $g$ by:

$g = (\tilde{h} - \min(\tilde{h}))/(\max(\tilde{h}) - \min(\tilde{h}))$, for a light figure on a dark background,

$g = (\max(\tilde{h}) - \tilde{h})/(\max(\tilde{h}) - \min(\tilde{h}))$, for a dark figure on a light background.

Let $\Sigma(A)$ denote a real number that is the sum of all cell values of an array $A$. Then one can choose the code length about $m \approx \alpha \Sigma(g)$, for a fixed $\alpha > 0$, in order to make the number of code points proportional to the image "foreground mass". One can note that the array $g/\Sigma(g)$ has the properties of a discrete probability function, however, certain cells have a zero value, with the consequence that certain cumulative probability functions are not strictly increasing, and thus they cannot be inversed. A simple solution to this problem consists of adding to each cell of the array $g$ a small positive quantity $c$ such as:

$$c = \lambda \, \Sigma(g)/\prod_{i=1}^{n} S_i,$$

where $\lambda$ is a small positive constant (e.g. $\lambda = 0.0001$). This limited "lighting of the background" has the same role as the strictly positive kernel functions used in the original method [1]. Finally, the discrete probability function from which we are going to build the density code is given by the array $f$ defined by:

$$f = (g + c)/\Sigma(g + c).$$

The remaining difficulty results from the fact that $f$ contains only a finite set of values, and thus the same is true for any cumulative function computed from $f$. As a consequence, one can find an infinite number of values of $U \in (0,1)^n$ for which there is no corresponding cell in $f$. This problem can be solved using a discrete dichotomic bounding search completed with a local linear interpolation. Given a value of $U \in (0,1)^n$, one can compute its corresponding code point $X = P^{-1}[f](U) \in R_h$ as follows:

```
function P⁻¹[f](u₁,...,uₙ) returns (x₁,...,xₙ)
fₙ ← f
for k ← n downto 1 do
    Pₖ[f](0) ← 0                              % computation of the vector Pₖ[f](0:Sₖ)
    for i ← 1 to Sₖ do  Pₖ[f](i) ← Pₖ[f](i−1) + Σ fₖ(i₁,...,i_{k−1},i)  endfor i
                                                i₁,...,i_{k−1}
    Pₖ[f](1:Sₖ) ← Pₖ[f](1:Sₖ)/Pₖ[f](Sₖ)
    inf ← 0,  sup ← Sₖ                        % dichotomic search
    while (sup − inf) > 1 do
        mid ← (inf + sup) div 2
        if uₖ ≥ Pₖ[f](mid) then inf ← mid else sup ← mid endif
    endwhile
    w ← (uₖ − Pₖ[f](inf))/(Pₖ[f](sup) − Pₖ[f](inf))        % linear interpolation
    xₖ ← inf + w sup
    if k > 1 then
        if inf > 0 then  f_{k−1} ← fₖ(1:S₁,...,1:S_{k−1},inf) + w fₖ(1:S₁,...,1:S_{k−1},sup)
                   else  f_{k−1} ← w fₖ(1:S₁,...,1:S_{k−1},sup)  endif
    endif
endfor k
```

In the above pseudo-code, the notation "$a:b$" is that of Matlab, and it refers to the index range $[a, a+1,...,b-1,b]$. Any text at right of "%" is a comment. Note that, for computational effectiveness, it is preferable to implement a specific version for each dimension $n$, as illustrated by the Matlab function "ImageCode", listed in the Appendix, for $n = 2$. One must also take care that the order of variables determines the set of coordinate transformations to which the comparison of codes can be made invariant. Typically, for image data, the order $(x, y)$ of the geometrical plane coordinates corresponds to a more probable variety of natural transformations than the reverse order $(y,x)$. However, the first dimension of an array usually corresponds to the $y$ coordinate, and the second dimension corresponds to the $x$ coordinate. So, the implementation must consider the dimensions in the most appropriate order, which is not necessarily the order of the array dimensions.

It remains to choose a sequence of points $(U_1,...,U_m)$ uniformly distributed in $(0,1)^n$, and to compute the code point corresponding to each of these points by the mapping $P^{-1}[f](U_j)$, $1 \le j \le m$, in the same order, which provides the desired density code. The length $m$ of the sequence can vary, if necessary, depending on the image size and complexity, however, for every given index $j$, the point $U_j$ must always be the same in order to make different density codes comparable. A good choice is to use a quasi-uniform sequence such as a Halton sequence [3] or a Faure sequence [4,5], as in [1]. It is well known that Faure sequences must be preferred for high dimension spaces, however, the data arrays considered in this paper have rarely more than three dimensions, thus one can as well use simple Halton sequences, and the Matlab function named "Halton" in the Appendix generates such sequences.

*Dissimilarity Function*—Given two density codes $V$ and $W$, that are $m \times n$ real matrices, both computed using the same quasi-uniform sequence, and given a chosen family $\Psi$ of transformation mappings, one can attempt to find a transformation $\tau \in \Psi$ that minimizes the quadratic matching error:

$$E_\Psi^2(V,W) = \min_{\tau \in \Psi} \|\tau(V) - W\|^2.$$

This minimization problem is very easy to solve if the transformation family $\Psi$ is linear in its parameters, which is the case, for example, of the family of multivariate polynomials of a given degree $d$, that has the special advantage of naturally including the family of affine transformations (first degree polynomials). In this case, one computes the polynomial basis functions for each point in $V$, resulting in a real matrix $B_V$ of order $m \times q$, where the number of monomials $q$ depends on $n$ and $d$. Then the $q \times n$ real matrix $T$ of the optimal polynomial coefficients is simply given by $T = B_V^\dagger W$, where $B_V^\dagger$ is the pseudo-inverse of $B_V$ [6,7]. In [1], it was proposed to use a dissimilarity function defined by:

$$\delta_\Psi(V,W) = \|B_V T - W\|/\sqrt{m}.$$

This dissimilarity function works well for data that conform to the basic probabilistic model, however, image functions do not behave exactly as probability functions, due to the presence of lighting variations, shadows, non-uniform background, and other sources of noise. As a result, when one compares two similar shapes, there is frequently a small proportion of code points that do not match and that provide very large errors. These points are outliers, and it is desirable to limit their effect on the dissimilarity measure. A simple solution to this problem is to replace the square root of the mean quadratic matching error, which is sensitive to outliers, by the median matching error, which is much less sensitive to outliers. Another difficulty, pointed out by [1], is that the dissimilarity measure is asymmetric and has the scale of the target code ($W$). This makes dissimilarity measures hard to compare when one works on a set of images that have different sizes. The solution is to make the dissimilarity measure relative to some evaluation of the scale of the target code. For example, one can divide the median matching error by the median distance of the target code points to their center of gravity (and multiply the result by 100 in order to obtain more readable numbers). Finally, in [1], only density codes having the same length were considered comparable, however, if two codes have different length, say $m_1$ and $m_2$, then the first $m = \min(m_1, m_2)$ code points are in fact comparable since they have been computed using the same quasi-uniform sub-sequence. Thus, we can compare density codes of different length, and we are no longer constrained to use a unique code length for all items in a database. The minor, yet useful, improvements of $\delta_\Psi$ suggested above are implemented in the Matlab function "DeltaMedian" listed in the Appendix.

## 3. Results

*Codes*—In order to test the performance of the above described method, we generated 6 pairs of images, each pair including a "plant-like" blurred fractal (A), and a "wind-like" transformation of it (B). Figure 1 shows the 12 test images, together with their first 1025 density code points generated using function calls of the form "*CodeName* = ImageCode (*DataArray*, U, 0)", with "U = Halton (1025, 2)" (see Appendix).

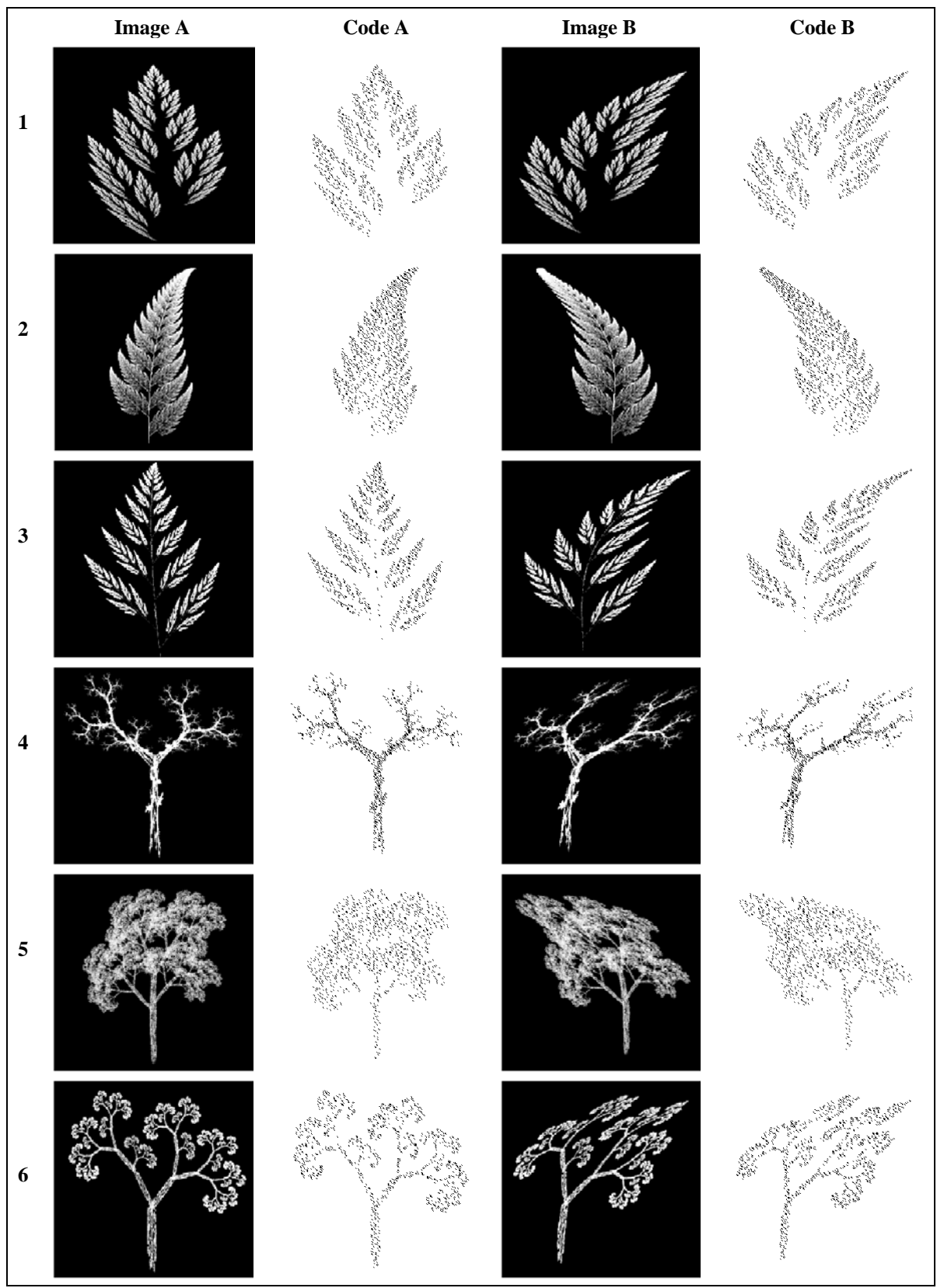

Figure 1. The 12 test images (256×256 pixels each) and their first 1025 density code points.

As one can see in Figure 1, the spatial distributions of code points suitably approximate the corresponding image foregrounds. Each code (1025 points for an image size of 256×256 pixels) was computed in about 19 milliseconds, in Matlab 7.4 running on a MacBook computer (Mac OS X, version 10.4.10), with a 2 GHz Intel Core 2 Duo processor.

*Dissimilarity Measure*—The "wind-like" transformations between A and B images of Figure 1 can be approximated by bivariate third-degree polynomial transformations, whose set was chosen as $\Psi$. The number of code points was made proportional (coefficient $\alpha$) to the image foreground mass, using function calls of the form "*CodeName* = ImageCode (*DataArray*, U, 0, $\alpha$)" (see Appendix), while $\alpha$ was experimentally varied from 0.01 to 0.5 (step 0.01). The foreground masses of test images ranged 3729-8923, and a long enough Halton sequence was available in all cases. Dissimilarity measures ($\delta_\Psi$) were computed for all pairs of distinct images, for the two possible argument orders (since $\delta_\Psi$ is asymmetric), and for all $\alpha$ values. The function calls were of the form "*Delta* = DeltaMedian (*Code1*, *Code2*, 3)" (see Appendix). For each $\alpha$ value, one selected the minimum and maximum obtained $\delta_\Psi$ values, for pairs of unrelated shapes (distinct "plants"), and for pairs of related shapes (A-B "wind-like" transforms). The result is plotted in Figure 2.

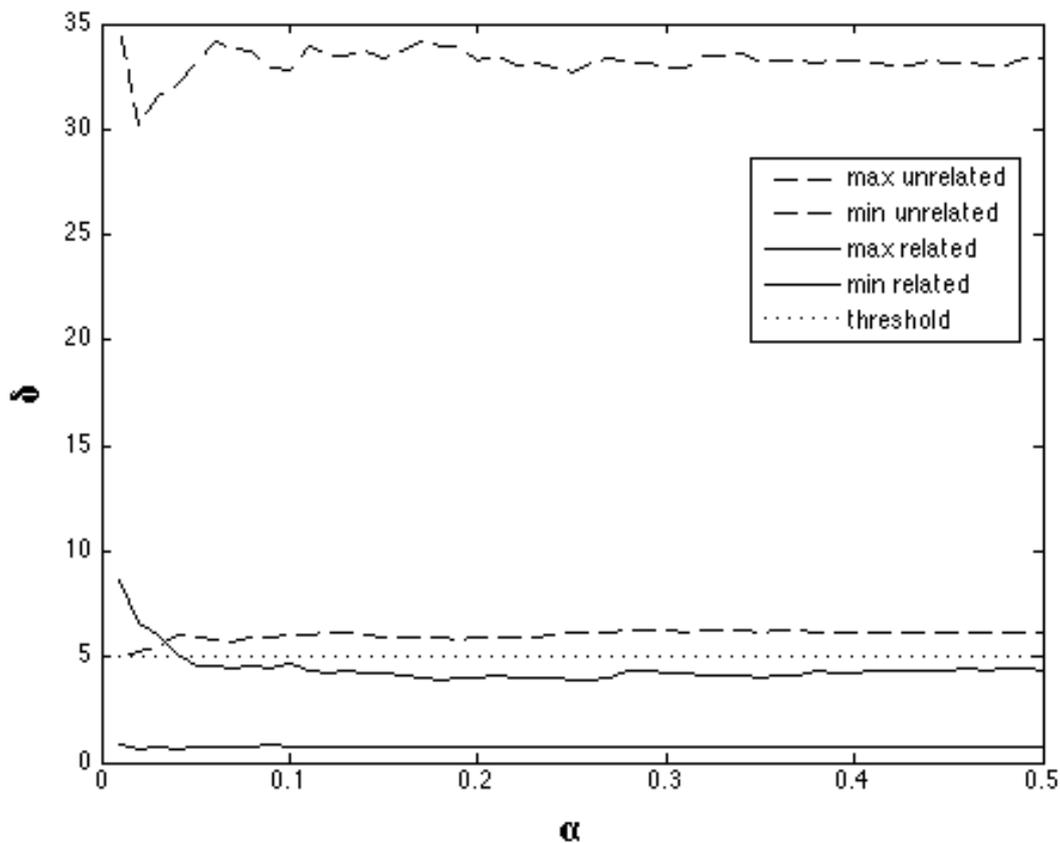

Figure 2. Observed $\delta_\Psi$ boundaries for related and unrelated image pairs, as functions of $\alpha$.

As one can see in Figure 2, for $\alpha > 0.04$, the related and unrelated $\delta_\Psi$ distributions do not overlap, the $\delta_\Psi$ values become quite stable, and one can reliably decide whether or not two images are related using a simple threshold of about 5. One can also note that the maximum separation of $\delta_\Psi$ distributions is reached for $\alpha =$ 0.25, however, this *a priori* depends on the considered shape space, and in particular on the relative scale of relevant distinctive features.

*Code Computation Time*—An examination of the operations involved in the image coding process shows that the complexity depends on both the image height ($H$), the image width ($W$), and the number of code points ($m$). The heaviest operations concern the image preprocessing ($O(HW)$), the $m$-times iterated dichotomic searches ($O(m(\log_2 H + \log_2 W - 2))$), and the $m$-times iterated computation of interpolated row vectors ($O(mW)$). In order to estimate the weights of these operations (for our platform), we measured the encoding time of random images whose height and width where independently varied from 16 to 1024 pixels (in powers of 2), while the code length was also independently varied from 16 to 1024, with repeated measures using 20 independent random images per condition. Then the regression equation was solved using a least square method, and we obtained the following approximation of the code computation time (in milliseconds):

$$t(H,W,m) \approx 10^{-4}(0.6853\,HW + 3.8459\,m(\log_2(HW)-2) + 0.3943\,mW),\quad (\pm 2.66).$$

The standard approximation error is small enough for practical use, and the correlation between the observed and approximated computation times is $r = 0.99$. As an example, the approximated computation time for the images of Figure 1, with 1025 code points, is $20.4 \pm 2.66$ milliseconds, whereas the observed computation time was about 19 milliseconds. We observed that adding more terms to the regression equation does not significantly improve the approximation accuracy, whereas the formula is obscured by the presence of negative coefficients.

Finally, we note that the obtained computation times make the density code approach perfectly usable for on-line computation and for the processing of large image databases. Using the original encoding algorithm, the code computation time, as reported in [1], was of 6 minutes and 45 seconds for 2048 code points on an image of 200×200 pixels. Even though the used computers are not the same, there is no doubt that the present algorithm considerably improves the situation, performing a similar encoding in only 29.4 milliseconds.

## Appendix

The following implementation code, in Matlab 7.4, is provided for example, and for academic use only. The code is not optimized and exception cases are not managed.

```
function code = ImageCode(f,u,DarkOnLight,alpha)
% Density code of an image f for a quasi-uniform sequence u
% For a fixed length code, do not provide the alpha argument
% Set DarkOnLight=1 for a dark figure on a light background (else 0)
[ymax,xmax]=size(f); minf=min(min(f)); maxf=max(max(f));
if DarkOnLight>0, f=(maxf-f)/(maxf-minf); else f=(f-minf)/(maxf-minf); end
Sf=sum(sum(f));
if nargin<4, m=length(u); else m=min(length(u),round(alpha*Sf)); end
lambda=0.0001; f=f+lambda*Sf/(ymax*xmax); Pyf=sum(f,2);
Pyf=cumsum(Pyf); Pyf=Pyf/Pyf(ymax);        % Pn[f] is computed only once
for p=1:m
    v=u(p,2); lob=1; upb=ymax;
    if v<=Pyf(1), w=v/Pyf(1); Pxf=f(1,:)*w;
    else while (upb-lob)>1
            y=round((lob+upb)/2);
             if Pyf(y)>v, upb=y; else lob=y; end
          end
      w=(v-Pyf(lob))/(Pyf(upb)-Pyf(lob));
      u(p,2)=lob+(upb-lob)*w;
      Pxf=f(lob,:)+(f(upb,:)-f(lob,:))*w;
    end
    Pxf=cumsum(Pxf); Pxf=Pxf/Pxf(xmax);   % Pk[f],k<n, is computed m times
    v=u(p,1); lob=1; upb=xmax;
    if v>Pxf(1)
        while (upb-lob)>1
            x=round((lob+upb)/2);
            if Pxf(x)>v, upb=x; else lob=x; end
        end
        u(p,1)=lob+(upb-lob)*(v-Pxf(lob))/(Pxf(upb)-Pxf(lob));
    end
end
code=u(1:m,:);

function u = Halton(m,n)
% Halton quasi-uniform sequence of m points in (0,1)^n
p=zeros(n,1);
```

```
p(1,1)=2;
for k=2:n
    p(k,1)=p(k-1,1)+1;
    while ~isprime(p(k,1)), p(k,1)=p(k,1)+1; end
end
u=zeros(m,n);
for t=1:m
    u(t,:)=point(n,t,p);
end

function pt=point(n,t,p)
pt=zeros(1,n);
for k=1:n
    pk=p(k,1); i=t; h=0; ib=1/pk;
    while (i>0)
        d=mod(i,pk);
        h=h+d*ib;
        i=round((i-d)/pk);
        ib=ib/pk;
    end
    pt(1,k)=h;
end

function delta = DeltaMedian(c1, c2, d)
% Delta function with d-degree polynomial invariants
% Modified from the DeltaPoly function listed in Courrieu (2006)
[m1,n]=size(c1); [m2,n]=size(c2); m=min(m1,m2);
c1=c1(1:m,:); c2=c2(1:m,:);      % reduce to comparable sub-sequences
if d == 0                        % direct comparison of density codes
    err = sqrt(sum((c1-c2).^2,2));
else                             % comparison of codes using invariants
    pw = AllPowers(n,d);
    [n,NbrTerms] = size(pw);
    x = ones(m,NbrTerms);
    for t = 1:NbrTerms
        for i = 1:n
            x(:,t) = x(:,t).*c1(:,i).^pw(i,t);
        end
    end
    T = pinv(x)*c2;              % optimal transformation coefficients
    err = sqrt(sum((x*T - c2).^2,2));
end
delta=median(err);               % median mismatch based dissimilarity
TargetCentre=mean(c2);
TargetScale= median(sqrt(sum((c2-kron(ones(m,1),TargetCentre)).^2,2)));
delta=100*delta/TargetScale;

function pwrs = AllPowers(n,d)
% All vectors of n positive integers of sum <= d
global PW;
PW = [];
for k = 0:d
    kPowers(n,[],k);
end
pwrs = PW;

function kPowers(n,v,k)
% Recursively builds vectors of sum k
global PW;
if length(v) == (n-1)
    v = [v;k];
    PW = [PW,v];
else
    for p = 0:k
        kPowers(n,[v;p],k-p);
```

254

```
        end
end
```

## References

[1] P. Courrieu, "Density Codes and Shape Spaces", *Neural Networks, Vol. 19*, pp. 429-445, 2006.
[2] D.F. Specht, "Probabilistic Neural Networks", *Neural Networks, Vol. 3*, pp. 109-118, 1990.
[3] J.H. Halton, "On the Efficiency of Certain Quasi-random Sequences of Points in Evaluating Multi-dimensional Integrals", *Numerische Mathematik, Vol. 2*, pp. 84-90, 1960.
[4] H. Faure, "Discrépance de Suites Associées à un Système de Numération (en Dimension s)", *Acta Arithmetica, XLI*, pp. 337-351, 1982.
[5] H. Faure, "Variations on (0, s)-sequences", *Journal of Complexity, Vol. 17*, pp. 741-753, 2001.
[6] A. Ben Israel, & T.N.E. Greville, *Generalized Inverse: Theory and Applications* ($2^{nd}$ ed.), New York, Springer, 2003.
[7] P. Courrieu, "Fast Computation of Moore-Penrose Inverse Matrices", *Neural Information Processing-Letters and Reviews, Vol. 8, No. 2*, pp. 25-29, 2005.

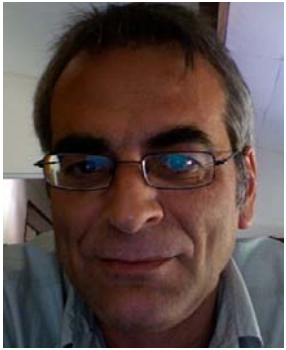

**Pierre Courrieu** received his PhD degree from University of Provence in 1983, and he is a CNRS researcher currently working with psychologists and neuroscientists in Marseille (France). He is a member of the European Neural Network Society, and his research interests include visual shape recognition, neural computation, data encoding, function approximation, supervised learning, and global optimization methods.